# Accelerating Returns and the Qualitative Engine for Science

Guojun Liao

Department of Mathematics, University of Texas at Arlington

**Abstract.** Ray Kurzweil's thesis [1, 2] of accelerating returns has been one of the most influential narratives in discussions of technological progress. Its central claim is that advances in multiple technological fields, especially compute, artificial intelligence, brain science, and biotechnology—interact so that progress becomes self-amplifying and approximately exponential. This paper offers a simple mathematical interpretation of that claim and then argues that, even if such acceleration is real, it does not by itself resolve the central problem of scientific discovery. The reason is that accelerating returns apply most naturally to executional and infrastructural capability, whereas genuine discovery often depends on a different capacity: qualitative reasoning about when a current framework is structurally inadequate and what conceptual move is needed next. Recent ARC-AGI-3 results sharpen this distinction by showing that the gap between current AI systems and human flexible reasoning remains very large. At the same time, Demis Hassabis has emphasized that humans must retain their sense of meaning and what they choose to focus their lives on, reminding us that the future of AI is not only a matter of technical prediction but also of what forms of human understanding are worth preserving and transmitting. This paper positions the Qualitative Engine for Science (QES) [3] as a response to that missing capacity. It also revisits the history of Chern's intrinsic proof of the Gauss–Bonnet theorem, not merely as a famous result, but as an example of framework lifting in mathematical discovery. In this view, Kurzweil's theory helps explain why quantitative capability may accelerate, while QES addresses the deeper problem that acceleration alone does not solve: how the highest forms of scientific judgment can be preserved, organized, and made accessible.

## 1. Introduction

Ray Kurzweil's theory of accelerating returns has shaped technological futurism for over two decades. In books such as The Singularity Is Near and The Singularity Is Nearer, the central idea is that progress in information technology does not proceed linearly. Instead, advances in one field become tools for accelerating advances in others, creating a positive-feedback system in which overall technological capability grows increasingly rapidly.

This picture has intuitive appeal. Better compute enables stronger artificial intelligence. Stronger artificial intelligence improves software engineering, chip design, and scientific data analysis. Improved biology and brain science suggest new architectures and new targets for AI systems. Advances in materials, fabrication, and data infrastructure feed back into hardware and algorithmic performance. Progress, in this view, becomes a coupled dynamical process rather than a collection of independent improvements.

The aim of this paper is not to reject that claim, but to interpret it more precisely and then identify its limit. The central argument is that Kurzweil's accelerating returns can be made mathematically coherent as a model of interacting technological subsystems. However, even if that model is substantially correct, it explains primarily the acceleration of executional capability rather than the qualitative formation and revision of scientific models. The latter remains a different problem.

That distinction motivates the role of the Qualitative Engine for Science (QES) [3]. QES is framed as a Layer 2 system: a qualitative reasoning engine trained on the processes of scientific discovery, especially on cases in which researchers identified structural inadequacy, missing conceptual companions, latent analogies, and framework replacements. The broader framework distinguishes three layers: Layer 1 search and summarization, Layer 2 qualitative reasoning, and Layer 3 quantitative execution. The stated target of QES is Layer 2.

This paper therefore develops a contrast. Kurzweil's theory helps explain why Layer 3 may accelerate. QES is designed for the part that remains: the recognition that the current framework itself may be wrong or incomplete. To sharpen this contrast, the paper revisits the history of Chern's intrinsic proof of the Gauss–Bonnet theorem and relates QES to Yoichi Ochiai's broader reflection that the AI era is not defined only by what machines can do, but also by what forms of human understanding we choose to preserve.

## 2. What Accelerating Returns Mean Mathematically

The phrase "accelerating returns" is often used rhetorically. A useful first step is to state what it means mathematically.

If a quantity X(t) grows exponentially, then its rate of change is proportional to its current size:

$$dX/dt = rX,$$

with solution

$$X(t) = X_0 e^{rt}$$

This does not mean infinite speed. It means that equal intervals of time produce equal percentage increases. The larger the quantity becomes, the larger its absolute increment becomes over the same time interval. In this sense, exponential growth is a signature of positive feedback.

For a single technological capability—say compute performance—this can be interpreted as follows: better compute helps design better chips, optimize better software, automate more engineering, and shorten the time to the next improvement. The output of the system contributes to improving the process that produces the next output.

Kurzweil's claim, however, is stronger than this. He does not describe one self-improving variable, but a system of multiple interacting technological domains.

## 3. Coupled Technological Growth

Let $X(t) = (C(t), A(t), B(t), M(t), \ldots)^T$ represent compute, AI capability, brain science knowledge, materials capability, and other relevant technological fields.

If each field evolved independently, one could write

$$\mathrm{dC/dt} = r_C\, C, \quad \mathrm{dA/dt} = r_A\, A, \quad \mathrm{dB/dt} = r_B\, B$$

But Kurzweil's picture is that these fields are coupled. A more realistic simplified model is

$$\mathrm{dX/dt} = \mathrm{MX}$$

where M is a matrix, whose diagonal terms represent intrinsic growth rates and whose off-diagonal terms represent cross-field contributions.

In the simplest two-field case,

$$\mathrm{dC/dt} = r_C\, C \; + \alpha \mathrm{A}$$

and

$$\mathrm{dA/dt} = r_A\, A + \beta \mathrm{C}$$

Here $r_C$ is compute's independent growth rate, $r_A\, A$ is AI's independent growth rate, α measures the contribution of AI to compute, and β measures the contribution of compute to AI.

The dominant growth rate of the coupled system is the larger eigenvalue of the matrix

$$[\, [r_C\, C, \alpha], [\beta, r_A\, A\,]\,]$$

namely

$$\lambda_+ = (r_C - r_A)/2 + \sqrt{(r_C - r_A)^2 + \alpha\beta}$$

This expression makes the logic clear. If $\alpha\beta > 0$, then the coupled system grows faster than either field would alone. In this precise sense, mutually reinforcing fields can produce accelerated, approximately exponential growth like $e^{\lambda_+ t}$.

The same logic extends to larger systems. Kurzweil's accelerating returns can therefore be interpreted as a coupled dynamical system in which progress in one field acts as an input to others, increasing the dominant eigenvalue of the whole technological system.

## 4. Conditions for Accelerating Returns

The mathematical form is simple, but it does not guarantee that accelerating returns will occur in practice. Several conditions are required.

4.1 Positive Feedback Must Be Real. The cross-coupling terms must correspond to genuine causal contributions. It is not enough that fields coexist. Progress in one domain must materially improve the rate of progress in another. Better AI must genuinely improve chip design, scientific modeling, engineering throughput, or experimental efficiency.

4.2 Transfer Friction Must Be Low. Even if useful insights exist in one field, they must be transferable. Shared representations, standard interfaces, accessible data formats, software ecosystems, and institutional structures all matter. If methods cannot move easily across domains, the effective coupling terms remain small.

4.3 Modularity and Recombination Must Be Available. The system must be modular enough so that improvements can be recombined. This is one reason software and digital infrastructure are so favorable to accelerating returns: output can often be copied, modified, and reused with relatively low cost.

4.4 Physical and Organizational Limits Remain. No technological subsystem grows exponentially forever. Energy, fabrication, coordination, regulation, scarce materials, and institutional bottlenecks impose limits. In many settings, exponential-looking growth is only a regime within a larger logistic or saturating curve. Thus, Kurzweil's picture is most plausible when viewed not as a timeless law, but as a regime description for highly coupled technical systems under favorable conditions.

## 5. What Kind of Capability Is Accelerating?

This is where the main argument of the present paper begins.

Even if one grants Kurzweil's model, an obvious question remains: what exactly is being accelerated? The answer is clearest in the language of the three-layer framework introduced in the earlier paper A Three-Layer Framework for AI in Scientific Discovery [3]. That paper argued that AI in scientific work can be separated into Layer 1: search and retrieval, Layer 2: model formation through qualitative reasoning, and Layer 3: execution, optimization, and refinement.

Kurzweil's mechanism maps most naturally onto Layer 3. It explains how executional power may accelerate: faster computation, larger models, more efficient code generation, better simulation, stronger optimization, and faster engineering loops. All of these fit naturally into feedback systems of the form $dX/dt = MX$.

But scientific discovery often requires something different: recognition that the current framework is structurally inadequate and that new variables, new constraints, or a new representational space are required. That is not merely stronger execution within a fixed model. It is a change in the model itself.

A system can therefore become exponentially stronger at execution and still remain trapped inside an inherited formulation. In that case, accelerating returns amplify capability without guaranteeing understanding.

## 6. The Role of QES

QES is motivated by two observations. First, that major scientific breakthroughs often require a qualitative leap—a conceptual or structural shift rather than a better quantitative optimization inside an existing framework. Second, current AI systems are strong in search and computation but weak in identifying dissatisfaction with a framework itself and proposing a new one.

The Qualitative Engine for Science is a subsystem designed precisely for the neglected Layer 2. The QES architecture places its main emphasis on Layer 2 and is trained on real examples of processes of scientific discoveries.

Each example is described in detail according to seven structural patterns built in QES, P1–P7, that characterize the qualitative structure of discovery: underdetermination, missing companion, latent analogy, golden particle, structural ripeness, multiple convergence, and framework lifting.

These patterns describe something that is fundamentally distinct from accelerating returns. They are not equations for capability growth. They are templates for moments in which a research program reaches a boundary and must identify what is missing and abandon the current framing altogether.

Thus, the contrast can be stated clearly: Kurzweil provides a theory of accelerating technical capability, which belongs Layer 3. QES provides a subsystem for Layer 2 which is aimed at qualitative structural and conceptual redirection. The two are not contradictory. In fact, they are increasingly complementary. The more Layer 3 becomes powerful and inexpensive, the more valuable correct Layer 2 direction becomes.

## 7. Why Acceleration Does Not Remove the Need for QES

One might object that sufficiently advanced AI, backed by accelerating returns, will eventually absorb qualitative reasoning as well. That is possible in principle, but it does not follow from the acceleration model itself.

The coupled-system model $dX/dt = MX$ describes the growth of capabilities within and across existing fields. It says nothing by itself about the ability to detect structural inadequacy, explain why a framework fails, or identify a neighboring domain that contains the missing conceptual object.

Those acts have different forms. They are closer to diagnosis of failure, recognition of missing structure, abductive inference, representational change, and temporal judgment about when a framework has exhausted itself.

These are exactly the capacities emphasized in the QES proposal. QES is contrasted with general-purpose AI systems that are strong at search and execution but weak at identifying which problem is worth solving next, which framework is incomplete, and which cross-domain connection is relevant.

This means that even in a world where Kurzweil's acceleration thesis is substantially correct, the need for a Layer 2 system remains. Indeed, it becomes sharper. A stronger execution engine pointed in the wrong direction simply reaches the dead end faster.

## 8. Chern's Intrinsic Gauss–Bonnet Proof as a Discovery Pattern

The history of Chern's intrinsic proof of the Gauss–Bonnet theorem, which connects topology to geometry deserves to be stated carefully. Chern did not create the problem ex nihilo.

Before his 1944 paper, important work had already been done by Hopf, Allendoerfer, Fenchel, and especially Allendoerfer–Weil. Hopf proved the hypersurface case; Allendoerfer and Fenchel generalized extrinsic curvature formulations to higher codimension in Euclidean space; and Allendoerfer–Weil gave a 1943 proof for the general case by reducing to local analytic embeddings. What remained unsatisfactory was that the theorem itself involved intrinsic invariants, while the proof still relied on embedding machinery.

The significance of Chern's 1944 work was therefore not that he was first to connect curvature and Euler characteristic in every sense, but that he gave the first truly intrinsic differential-form proof for closed even-dimensional Riemannian manifolds. After André Weil expressed the belief that intrinsic proof should exist, Chern found one within a short time. His proof replaced the dependence on extrinsic embedding by a transgression argument on the unit sphere bundle, constructed from connection and curvature forms. In modern language, this result stands at the beginning of what later became Chern–Weil theory.

This historical revisit matters for the present paper because it makes the discovery pattern more precise. The breakthrough was not a vague intuitive leap. It involved a clear qualitative transition: recognition of a structural limitation in earlier formulations, identification of the right intrinsic companion object, and reformulation of the problem in a broader conceptual language in which the theorem became natural. That is exactly the kind of framework lifting and missing companion reasoning that QES is designed to preserve, organize and systemize.

### 9. Relationship with the Earlier Three-Layer Paper

The earlier paper [3] argued that Layer 2 is the bottleneck in AI for scientific discovery and illustrated this with case studies such as Chern's intrinsic proof of Gauss–Bonnet, Ryu–Jang's Lyapunov analysis of Nesterov acceleration, and the OpenAI unit-distance result. Its purpose was to establish the existence and importance of Layer 2 as a distinct object of study.

The present paper has a different purpose. It does not attempt to restate the full case for the three-layer framework. Instead, it addresses a likely objection from technological acceleration narratives: if the whole AI system is accelerating, why is a specialized Layer 2 engine needed at all?

The answer offered here is that Kurzweil's theory and QES operate at different explanatory levels. Kurzweil explains how technical capability may accelerate through interaction. QES addresses the qualitatively different question of how frameworks are judged, lifted, and redirected. In this sense, the present note should be read as a complement to the earlier paper, not as a repetition of it.

### 10. The Enduring Value of QES Beyond AGI Timelines

A further reason for QES does not depend on any particular AGI timeline. Recent benchmark results such as ARC-AGI-3 [4] are useful because they show that the gap between current AI systems and human flexible reasoning remains large. They therefore support the practical case for intermediate human–AI systems. But there is a deeper point. The value of QES does not ultimately rest on whether AGI arrives in two years, five years, or later. It rests on a more stable fact: the processes by which scientific discoveries occur—how researchers recognize the boundary of a framework, how they identify what is structurally missing, how they search a broader conceptual neighborhood for the missing companion, and how they decide when a framework must be lifted—are themselves a form of human wisdom worth preserving, organizing, and transmitting.

This point is consistent with Demis Hassabis's insistence that humans must continue to maintain their sense of meaning and decide what they choose to focus their lives on [5]. That is not a technical forecast but a value judgment. It implies that the future role of AI is not exhausted by replacing human labor or matching benchmark scores. Even in a world of very powerful AI, there remains a distinct human question: what kinds of understanding, judgment, and intellectual inheritance should be retained and cultivated? QES should be understood in this light. It is not merely a temporary bridge before AGI. It is a human–AI interface designed to preserve and make accessible a body of qualitative scientific reasoning that might otherwise remain scattered, private, or historically buried.

There is also a second distinction. Even if AGI is interpreted operationally as broad competence across a large fraction of ordinary human cognitive tasks, scientific discovery often demands something rarer: not average human intelligence, but exceptional conceptual judgment. The major breakthroughs of science and mathematics are typically produced not by broad competence alone, but by the ability to detect structural inadequacy, make a conceptual reversal, and introduce a better framework. In that sense, even a future AI system with broad human-level general intelligence would not automatically replicate the kind of reasoning associated with the top tier of scientific creativity. QES addresses this by focusing not on generic intelligence in the abstract, but on the specific structures by which discovery proceeds.

For this reason, QES should not be justified only as a stopgap before AGI. Its deeper justification is that the understanding of scientific discovery processes is itself an enduring intellectual resource. Whether AGI arrives in 2030 or much later, that resource remains worth preserving and transmitting.

## 11. QES and Yoichi Ochiai's Perspective

A further way to understand the significance of QES is through the broader perspective developed by Yoichi Ochiai. Ochiai's work on Digital Nature [8, 9, 10] suggests that the AI era is not only about stronger models and more pervasive computation. It is also about how humans understand themselves, their relation to technology, and the conditions of meaning under a digitally transformed environment.

In this perspective, AI is not merely an instrument for replacing labor. It also functions as a new "other" through which humans reflect on what should remain distinctly human. The important question is therefore not only what machines can do, but what forms of human understanding are worth retaining, cultivating, and transmitting.

QES belongs naturally within that broader perspective. If Ochiai asks how AI changes human self-understanding in the age of Digital Nature, QES asks how the most valuable forms of scientific human judgment can be preserved and made accessible within that new condition. This connection is important because it prevents QES from being misunderstood as merely a stopgap before AGI. Even if AGI arrives soon, the question remains: what intellectual inheritance do we wish to preserve? In science, one answer is clear. We should preserve not only results, but also the qualitative pathways by which major discoveries become possible.

## 12. Conclusion

Kurzweil's accelerating returns can be given a clear mathematical interpretation through coupled dynamical systems with positive cross-field feedback. In that sense, the theory is coherent and captures something real about the interaction of compute, AI, and neighboring technological domains.

But the theory explains primarily the acceleration of executional capability. It does not by itself explain how systems identify structural inadequacy in a scientific framework, find a missing conceptual companion, or decide when a framework should be abandoned.

That is the role of the Qualitative Engine for Science. If Kurzweil is right, then Layer 3 may accelerate dramatically. But that does not eliminate the Layer 2 problem. It makes it more central.

Seen in this light, the revised history of Chern's intrinsic Gauss–Bonnet proof and the relation to Ochiai's Digital Nature perspective point in the same direction: the AI age is not defined only by what machines can do, but by what forms of human understanding we choose to keep alive.

## Acknowledgments

The ideas and structural insights in this paper were shaped by the author's research experience in mathematics and medical image analysis, supported in part by the National Science Foundation and the National Institutes of Health. The author also acknowledges the use of ChatGPT (OpenAI) as an AI assistant in literature search, drafting, and manuscript preparation.